\documentclass[letterpaper]{article} 
\usepackage{aaai2026}  
\usepackage{times}  
\usepackage{helvet}  
\usepackage{courier}  
\usepackage[hyphens]{url}  
\usepackage{graphicx} 
\usepackage{natbib}  
\usepackage{caption} 
\usepackage{algorithm}
\usepackage{algorithmic}
\usepackage{siunitx}

\usepackage{newfloat}
\usepackage{listings}
\DeclareCaptionStyle{ruled}{labelfont=normalfont,labelsep=colon,strut=off} 
\floatstyle{ruled}
\newfloat{listing}{tb}{lst}{}
\floatname{listing}{Listing}
\usepackage{multirow}
\usepackage{makecell}
\usepackage{amssymb}
\usepackage{amsmath}
\usepackage{booktabs}

\title{Planning Under Observation Mismatch for Traffic Signal Control\\via Adaptive Modular World Models}
\author {
    Zherui Huang,
    Yicheng Liu,
    Chumeng Liang,
    Guanjie Zheng\thanks{Corresponding author.}
}
\affiliations {
    Shanghai Jiao Tong University\\
    \{huangzherui, liuyicheng1515\}@sjtu.edu.cn, caradryan2022@gmail.com, gjzheng@sjtu.edu.cn
}

\begin{document}

\maketitle

\begin{abstract}
Deploying learned decision-making systems often requires transferring to new sites where the sensing pipeline differs. In such cases, observations can change in semantics and dimensionality even when action primitives and objectives remain comparable. In this work, we study transferable model-based planning under this observation mismatch, which remains challenging for existing learning-based approaches. We propose Adaptive Modularized Model (AMM), a modular planning architecture that separates a domain-specific observation adapter from a shared internal dynamics model defined in a common planning state space. The dynamics model is meta-learned from multiple source domains to enable fast adaptation with limited target interaction. At run time, AMM performs receding-horizon planning by rolling out candidate action sequences under the learned dynamics and selecting actions that optimize a task-specific objective over predicted futures. We instantiate the approach on cross-domain traffic signal control, where actions correspond to signal phases and the planning objective captures congestion. Experiments show that AMM improves both performance and data efficiency compared with existing conventional controllers and learning-based baselines.
\end{abstract}

\begin{links}
    \link{Code}{github.com/ZheruiHuang/AMMforTSC}
    \link{Datasets}{traffic-signal-control.github.io/index.html}
\end{links}

\section{Introduction}
\label{sec:intro}

Automated planning and scheduling systems are increasingly deployed in settings where the control problem repeats across many sites~\cite{ghallab2016automated}.
A recurring obstacle is that the sensing and logging pipeline changes from site to site.
Sensors differ in modality and coverage, while feature definitions differ due to environmental factors (e.g., vendors and jurisdictions)~\cite{xing2021domain}.
As a result, the observation space can change in semantics and dimension even when the available actions and the underlying physical process remain similar~\cite{sun2022transferobs}.
This form of observation mismatch makes it difficult to reuse learned decision-making components~\cite{zhao2020sim}.
It also raises the cost of deployment, since collecting extensive online interaction data at each new site can be expensive and risky~\cite{garcia2015comprehensive, levine2020offline}.

Model-based planning offers a natural solution to this challenge~\cite{mayne2000constrained, chua2018deep}.
If one can learn a compact planning state and a predictive model that captures the system dynamics, then the controller can evaluate candidate action sequences by rollout and select actions with a receding-horizon plan~\cite{rawlings2017model}.
Learning can supply the required components by fitting a state abstraction that exposes the variables needed for planning and a dynamics model that is accurate in the state space that matters for control~\cite{ha2018world, hafner2019learning}.
Recent work on world model planning has shown that decomposing a controller into a representation model, a dynamics model, and a value or prediction model can support strong performance in complex domains~\cite{hafner2019learning, hafner2019dream, schrittwieser2020mastering}.
However, most existing approaches are typically studied with a fixed observation interface and have not been the primary focus for transfer across changing observation pipelines.

This paper studies transferable planning under observation mismatch.
We consider a family of related domains that share the same action space and planning objective, and whose latent dynamics are similar, but whose observations differ.
Our goal is to obtain a planner for a new target domain using limited target interaction, while leveraging data collected from multiple source domains.
The key design choice is to separate what must change with the sensors from what should remain stable across domains.
We therefore introduce a modular architecture that decouples a domain-specific observation adapter from a shared internal dynamics model that supports lookahead planning.

We present Adaptive Modularized Model (AMM), an adaptive modular world model planner.
AMM maps raw observations to a shared planning state using a domain-specific representation module.
It then predicts the transition of that planning state under candidate action sequences using a shared dynamics module.
A value module scores predicted future states, which enables receding horizon planning by selecting the action sequence with the best predicted outcome.
The shared dynamics module is learned with meta learning across source domains so that it can take advantage of large amounts of multi-source heterogeneous data and be adapted to a new target domain with limited data and few updates.
Only the observation adapter must be replaced or retrained when the observation interface changes, which enables data-efficient reuse of the planning model across deployments with heterogeneous sensing.

In this work, we instantiate this framework in traffic signal control problems.
Traffic signal control can be viewed as a sequential planning problem~\cite{varaiya2013max, wei2019survey}.
At each control interval, each intersection selects a signal phase, and the joint decisions across intersections shape network-wide congestion over time.
Traffic systems also exhibit the deployment conditions that motivate our setting~\cite{zhang2019cityflow}.
Different cities and districts often provide different sensing and logging capabilities~\cite{zang2020metalight}.
Some provide only coarse counts, while others provide speeds or trajectory aggregates.
At the same time, the action primitives and the underlying queueing dynamics have a common structure across deployments~\cite{sun2024crosslight}.
These properties make cross-domain traffic signal control a suitable testbed for studying learning components that enable planning under observation mismatch \cite{zhang2019cityflow, wei2019survey, jiang2024x}.

We evaluate AMM on three public benchmarks with heterogeneous observations and limited target interaction.
Across all road networks, AMM improves both average travel time and average queue length compared with conventional controllers and strong learning-based baselines.
We further conduct ablations that isolate the effects of modularization and meta-learning, as well as analyses on model capacity, source-domain choice, and a case study that trains the observation adapter from logged data.

This work makes three main contributions.
First, we formalize transferable planning under observation mismatch, where observation spaces vary across domains while action semantics and objectives are shared.
Second, we propose AMM, a modular world model planner that combines a domain-specific observation adapter with a shared meta-learned dynamics model for planning.
Third, we evaluate AMM on traffic signal control problems and provide ablations that isolate the effects of modularization, meta-learning, and data budget on target-domain adaptation.

The remainder of this paper is organized as follows.
We first review related work on learned planning models, transfer under observation mismatch, and traffic signal control.
We then present AMM, including the modular world model, the meta-learning procedure, and the receding-horizon planner.
Next, we report experimental results and analyses.
Finally, we conclude with limitations and directions for future work.

\section{Related Work}
\label{sec:related_work}

\paragraph{Planning with learned models}
A central theme in learning for planning is to acquire predictive models and abstractions from data, and then use them for online decision-making.
Model-based reinforcement learning has explored planning by rollout with learned dynamics, including probabilistic ensembles and trajectory optimization~\cite{chua2018deep}, as well as latent world models that enable imagination-based control~\cite{hafner2019dream, hafner2019learning}.
MuZero popularized a representation, dynamics, and prediction decomposition and combined it with search~\cite{schrittwieser2020mastering}.
Our approach is aligned with this family of methods in its use of a learned latent dynamics model to support lookahead.
The main difference is the target setting.
We focus on transfer across domains with heterogeneous observations, and we structure the model so that observation handling is domain-specific while the internal dynamics used for planning are shared and adapted with limited target updates.

\paragraph{Transfer and adaptation under observation mismatch}
Transfer in reinforcement learning is often studied through domain randomization and augmentation~\cite{tobin2017domain}, as well as representation learning methods that aim to extract task-relevant factors that generalize across domains.
Meta learning provides a complementary approach by learning initial parameters that can be adapted quickly to new tasks~\cite{finn2017model}.
Several works address adaptation when the observation interface changes by learning representations that align across domains or by learning latent states that preserve controllable structure.
These methods motivate the separation of perception from dynamics and decision making.
Our work follows this principle, but it enforces it at the level of planning by learning a shared dynamics model in a common planning state space, while allowing the observation adapter to vary with the sensing pipeline.

\paragraph{Traffic signal control}
Traffic signal control has a long history in transportation research.
Classical methods include fixed time control~\cite{allsop1971delay}, self-organizing traffic lights~\cite{gershenson2004self}, and MaxPressure control~\cite{varaiya2013max}.
Deep reinforcement learning has been applied extensively to TSC and has produced strong results in simulation.
Representative methods include IntelliLight~\cite{wei2018intellilight}, FRAP~\cite{zheng2019learning}, PressLight~\cite{wei2019presslight}, CoLight~\cite{wei2019colight}, MPLight~\cite{chen2020toward}, AttendLight~\cite{oroojlooy2020attendlight}, MetaLight~\cite{zang2020metalight}, and UniLight~\cite{jiang2022multi}.
Recent work has also emphasized improved state representations for pressure-based control~\cite{wu2021efficient, zhang2022expression}.
CityFlow provides a widely used open simulator and benchmark suite for large-scale evaluation~\cite{zhang2019cityflow}.
LibSignal further standardizes datasets, environments, and implementations for more reproducible comparisons~\cite{mei2024libsignal}.

\paragraph{Cross-domain and data-efficient TSC}
The cost of online exploration has motivated research on cross-domain transfer and offline to online learning in TSC.
Recent cross-city methods leverage offline data and limited target interaction to improve deployability~\cite{sun2024crosslight, jiang2024x}.
Other lines study offline reinforcement learning for TSC~\cite{zhang2023datalight, bokade2025offlight} and practical constraints such as safety and missing or partial observations~\cite{du2023safelight, mei2023reinforcement}.
These works support the importance of transfer and limited data adaptation in realistic deployments.
Most existing approaches focus on transferring policies or coordinating representations under a fixed observation interface.
In contrast, our method explicitly targets observation function mismatch by modularizing the observation adapter and transferring the internal dynamics used for lookahead planning.

\section{Method}
\label{sec:method}

\subsection{Problem Formulation}
\label{sec:formulation}

We study transferable model-based planning across a set of related domains.
Each domain $d \in \mathcal{D}$ is a controlled process with a shared action space $\mathcal{A}$ and an objective that is comparable across domains.
At time $t$, the agent receives an observation $o_t \in \mathcal{O}_d$ and selects an action $a_t \in \mathcal{A}$.
The observation space $\mathcal{O}_d$ may differ in semantics and dimensionality across domains.
This captures changes in sensing and preprocessing pipelines.

We assume there exists a shared \emph{planning state space} $\mathcal{S}$ that is sufficient for decision making.
The planning state is not directly observed.
Instead, each domain provides observations through a domain-specific observation process.
Our goal is to obtain a planner for a target domain $d^\star$ using limited target interaction, while leveraging data from a set of source domains $\mathcal{D}_S$.

AMM addresses this setting by learning
(i) a domain-specific observation adapter that maps observations into $\mathcal{S}$, and
(ii) a shared internal dynamics model in $\mathcal{S}$ that supports online lookahead.
At deployment time, AMM performs receding-horizon planning by rolling out candidate action sequences using the learned dynamics and selecting actions that optimize a planning objective over predicted futures.

\begin{figure}
    \centering
    \includegraphics[width=\linewidth]{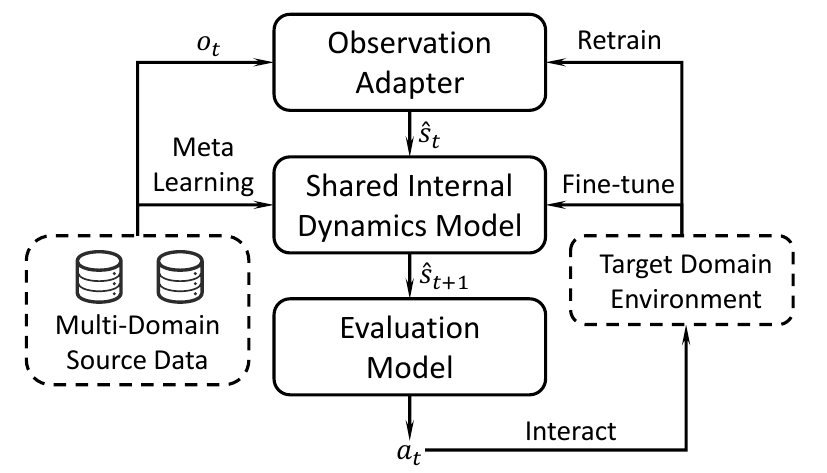}
    \caption{Framework of Adaptive Modularized Model}
    \label{fig:overview}
\end{figure}

\subsection{Adaptive Modular World Model Planning}
\label{sec:amm_overview}

AMM separates components that depend on the observation interface from components that should transfer across domains.
It contains three parts.
The first part is an observation adapter.
The second part is an internal dynamics model.
The third part is an evaluation model used for planning.
The framework of AMM is illustrated in Figure~\ref{fig:overview}.

\subsubsection{Observation Adapter}
\label{sec:adapter_general}

The observation adapter is domain-specific.
It maps a domain observation to a shared planning state representation
\begin{equation}
\hat{s}_t = f_{\phi_d}(o_t), \qquad \hat{s}_t \in \mathcal{S}.
\label{eq:adapter_general}
\end{equation}
The parameters $\phi_d$ are allowed to vary across domains to accommodate heterogeneous observation spaces.
This is the only component that must be replaced or retrained when the observation interface changes.

\subsubsection{Shared Internal Dynamics Model}
\label{sec:dynamics_general}

The internal dynamics model is shared across domains.
It predicts how the planning state transitions under actions as
\begin{equation}
\hat{s}_{t+1} = g_{\theta}(\hat{s}_t, a_t).
\label{eq:dynamics_general_1step}
\end{equation}
For a planning horizon $H$, we roll out iteratively
\begin{equation}
\hat{s}_{t+h+1} = g_{\theta}(\hat{s}_{t+h}, a_{t+h}), \quad h = 0, \dots, H-1.
\label{eq:dynamics_general_rollout}
\end{equation}
The parameters $\theta$ are shared across all source domains and the target domain.
They are learned from source-domain data and adapted in the target domain with few updates.

\subsubsection{Evaluation Model and Receding-Horizon Planning}
\label{sec:planning_general}

AMM uses an evaluation model to score predicted futures.
We denote the evaluation function by $V: \mathcal{S} \to \mathbb{R}$.
It can be a learned value model or an explicit objective proxy.
Given a candidate action sequence $\mathbf{a}_{t:t+H-1} = (a_t, \dots, a_{t+H-1})$, AMM rolls out \eqref{eq:dynamics_general_rollout} and computes
\begin{equation}
\mathrm{Score}(\mathbf{a}_{t:t+H-1})
= \sum_{h=1}^{H} \gamma^{h-1} V(\hat{s}_{t+h}),
\label{eq:score_general}
\end{equation}
where $\gamma \in (0,1]$ discounts future outcomes within the horizon.
AMM selects the sequence with the highest score and executes the first action
\begin{equation}
\mathbf{a}_{t:t+H-1}^{\star}
= \arg\max_{\mathbf{a}_{t:t+H-1} \in \mathcal{A}^H} \mathrm{Score}(\mathbf{a}_{t:t+H-1}),
\quad
a_t \leftarrow \mathbf{a}_{t}^{\star}.
\label{eq:rhp_general}
\end{equation}
This produces receding-horizon planning by rollout in a learned planning state space.

\subsection{Learning and Adaptation}
\label{sec:learning}

AMM learns the domain adapters $\{f_{\phi_d}\}$ and a transferable initialization of the shared dynamics parameters $\theta$.

\subsubsection{Losses}
\label{sec:losses}

We train the dynamics model with multi-step prediction loss in the planning state space.
For a trajectory segment starting at time $t$, let $\hat{s}_{t+h}$ be the rollout produced by \eqref{eq:dynamics_general_rollout}.
We minimize
\begin{equation}
\mathcal{L}^{\mathrm{dyn}}_d(\theta)
= \mathbb{E}\Big[ \sum_{h=1}^{H} \lambda^{h-1} \, \mathrm{Dist}(\hat{s}_{t+h}, s_{t+h}) \Big],
\label{eq:dyn_loss_general}
\end{equation}
where $\lambda \in (0,1]$ discounts longer rollouts and $\mathrm{Dist}(\cdot,\cdot)$ is a task-dependent distance on planning states.

In our instantiation, the planning state $s_t$ can be computed from the simulator, which allows supervised training of the observation adapter.
We minimize
\begin{equation}
\mathcal{L}^{\mathrm{repr}}_d(\phi_d)
= \mathbb{E}\Big[ \mathrm{Dist}\big(f_{\phi_d}(o_t), s_t\big) \Big].
\label{eq:repr_loss_general}
\end{equation}

\subsubsection{Meta-Learning the Dynamics Initialization}
\label{sec:maml}

We treat each source domain as a task and meta-learn an initialization for $\theta$ so that it adapts quickly to a new domain.
We use model-agnostic meta-learning~\cite{finn2017model}.
At each meta-iteration, we sample a meta-batch of domains $\mathcal{B} \subset \mathcal{D}_S$.
For each $d \in \mathcal{B}$, we split its data into a support set $\mathcal{S}_d$ and a query set $\mathcal{Q}_d$.
We compute task-adapted parameters by a gradient step on the support loss
\begin{equation}
\theta'_d = \theta - \alpha \nabla_{\theta}\mathcal{L}^{\mathrm{dyn}}_{d}(\theta; \mathcal{S}_d),
\label{eq:maml_inner_general}
\end{equation}
then update the meta-parameters by minimizing the query losses after adaptation
\begin{equation}
\theta \leftarrow \theta - \beta \nabla_{\theta} \sum_{d \in \mathcal{B}}
\mathcal{L}^{\mathrm{dyn}}_{d}(\theta'_d; \mathcal{Q}_d).
\label{eq:maml_outer_general}
\end{equation}
The sum in \eqref{eq:maml_outer_general} is over domains in the meta-batch.
It is not a sum over minibatches within a single domain.
The meta-learning procedure is shown in Algorithm~\ref{alg:maml_dyn}.

\begin{algorithm}[t]
\caption{Meta-learning the dynamics initialization}
\label{alg:maml_dyn}
\begin{algorithmic}[1]
\REQUIRE Source domains $\mathcal{D}_S$, step sizes $\alpha, \beta$
\STATE Initialize dynamics parameters $\theta$
\WHILE{not converged}
\STATE Sample meta-batch $\mathcal{B} \subset \mathcal{D}_S$
\FOR{each domain $d \in \mathcal{B}$}
\STATE Sample support set $\mathcal{S}_d$ and query set $\mathcal{Q}_d$
\STATE $\theta'_d = \theta - \alpha \nabla_{\theta}\mathcal{L}^{\mathrm{dyn}}_{d}(\theta; \mathcal{S}_d)$
\ENDFOR
\STATE $\theta \leftarrow \theta - \beta \nabla_{\theta} \sum_{d \in \mathcal{B}} \mathcal{L}^{\mathrm{dyn}}_{d}(\theta'_d; \mathcal{Q}_d)$
\ENDWHILE
\RETURN $\theta$
\end{algorithmic}
\end{algorithm}

\subsubsection{Target-Domain Adaptation and Deployment}
\label{sec:adaptation}

Given a target domain $d^\star$, we initialize the dynamics model with the meta-learned parameters $\theta$.
We then learn the target adapter $f_{\phi_{d^\star}}$ and fine-tune $\theta$ using a limited interaction budget in $d^\star$.
After adaptation, AMM controls the target domain by applying \eqref{eq:adapter_general} and \eqref{eq:rhp_general} at each decision step.
The procedure of target-domain adaptation is shown in Algorithm~\ref{alg:amm}.

\begin{algorithm}[t]
\caption{AMM adaptation and control in the target domain}
\label{alg:amm}
\begin{algorithmic}[1]
\REQUIRE Source domains $\mathcal{D}_S$, target domain $d^\star$
\REQUIRE Horizons $H$, step sizes $\alpha,\beta,\eta$
\STATE Meta-learn $\theta$ using Algorithm~\ref{alg:maml_dyn}
\STATE Initialize target adapter parameters $\phi_{d^\star}$
\WHILE{target interaction budget not exhausted}
\STATE Collect target data and update $\phi_{d^\star}$ by minimizing $\mathcal{L}^{\mathrm{repr}}_{d^\star}(\phi_{d^\star})$
\STATE Fine-tune $\theta$ by minimizing $\mathcal{L}^{\mathrm{dyn}}_{d^\star}(\theta)$
\ENDWHILE
\STATE Deploy with receding-horizon planning using \eqref{eq:rhp_general}
\end{algorithmic}
\end{algorithm}

\subsection{Instantiation to Traffic Signal Control}
\label{sec:tsc_instantiation}

\begin{figure}
    \centering
    \includegraphics[width=\linewidth]{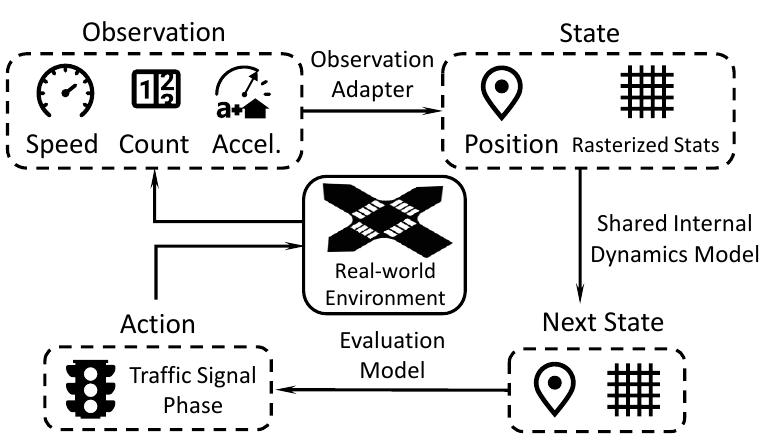}
    \caption{Illustration of instantiation to traffic signal control}
    \label{fig:instantiation}
\end{figure}

We now specify how the formulation and modules above are instantiated for traffic signal control.

\subsubsection{Domain and actions}
A domain corresponds to a road network and traffic flow configuration with a particular observation interface.
The road network contains $M$ signalized intersections.
At each control interval $t$, intersection $i$ selects an action $a_t^i$ from a discrete phase set $\mathcal{A}^i$.
AMM uses parameter sharing across intersections.
It applies the same model parameters to every intersection and outputs one action per intersection at each step.
Planning is performed per intersection using \eqref{eq:rhp_general}, and it can be executed in parallel across intersections.

\subsubsection{Planning state}
For each intersection $i$, the planning state $s_t^i \in S_i$ lies in a fixed domain-invariant space.
In this work, we consider two components of $s_t^i$:
(1) normalized vehicle positions near the stop line for each incoming lane, and
(2) rasterized approach-level statistics over segments, including estimated vehicle counts and average speeds.

\subsubsection{Observation mismatch}
Each domain $d$ exposes an observation vector $o_t^{i,d}$ for each intersection.
The observation components differ across domains in dimension and semantics, such as counts, speeds, or flow aggregates.
The adapter $f_{\phi_d}$ maps these heterogeneous observations to $\hat{s}_t^i$, so that the shared dynamics model $g_{\theta}$ operates in a consistent planning state space across domains.

\subsubsection{Evaluation function}
We score predicted futures using a congestion-oriented value proxy.
We aggregate occupancy in $n$-cell segments and apply larger weight to segments closer to the stop line.
This defines $V(s)$ and $\mathrm{Dist}(\cdot,\cdot)$ used in \eqref{eq:score_general}, \eqref{eq:dyn_loss_general}, and \eqref{eq:repr_loss_general}.
Details are provided in the experimental section and match the objective of reducing congestion and queueing near intersections.

\section{Experiments}
\label{sec:experiments}

\subsection{Experimental Setup}
\label{sec:exp_setup}

\subsubsection{Simulator and Benchmarks}
We use CityFlow \cite{zhang2019cityflow}, an open-source microscopic traffic simulator designed for large-scale traffic signal control research.
CityFlow provides efficient simulation of vehicle movements on real road networks and has been widely used in recent TSC benchmarks.

We evaluate on three public benchmarks derived from real traffic data and road networks \cite{zhang2019cityflow,wei2019survey}.
The benchmarks are Hangzhou ($4\times4$), Manhattan ($16\times3$), and Manhattan ($28\times7$), where $x\times y$ denotes the network size, i.e., the number of signalized intersections is roughly arranged in a grid with $x$ rows and $y$ columns.
Each benchmark includes a road network file and a vehicle flow file of 3600 seconds.
We treat each benchmark as a separate domain.
Although two benchmarks are from Manhattan, they correspond to different network sizes and traffic patterns, and they constitute distinct domains in our evaluation.

\subsubsection{Observation, State, Value, and Action}
To study observation-function shift, we assign each domain a different observation vector.
All domains include a shared core feature consisting of per-lane vehicle counts near the intersection.
We then add a domain-specific feature to induce mismatch.
Hangzhou ($4\times 4$) provides the number of vehicles that entered the intersection during the last control interval.
Manhattan ($16\times 3$) provides the number of vehicles passing the midpoint of each road segment during the last control interval.
Manhattan ($28\times 7$) provides average speeds measured in two road segments, namely the middle third and the last third.
AMM uses only these observations as input to its adapter.
For baselines, we use the default observation definitions from their implementations, since these methods are not designed for varying observation interfaces across domains.

The internal planning state is domain‑invariant and is constructed from simulator ground truth for training and evaluation.
It has two complementary parts.
First, for each incoming lane, we represent every vehicle by its distance‑to‑stop‑line expressed as a percentage of the lane length.
Second, we add a rasterized summary over each approach.
For every 10\,m segment, we record estimated vehicle count, average speed, and related statistics.
The combination captures microscopic proximity and mesoscopic flow and is sufficient to support effective planning in our experiments.
These choices are representative rather than mandatory.
Other encodings can replace them without changing the method.

We use the average queue length to quantify congestion.
It is computed over incoming lanes as the mean number of standing vehicles near the stop line.
Lower values indicate better traffic conditions and serve as our planning objective proxy and evaluation metric.

At each control interval, each intersection selects one of eight standard signal phases.
The chosen phase is held until the next decision step.

\begin{table*}
    \centering
    \setlength{\tabcolsep}{3.5pt}
    \renewcommand{\arraystretch}{1.05}
    \begin{tabular}{
    l
    S[table-format=4.2] @{$\,\pm\,$} r 
    S[table-format=3.2] @{$\,\pm\,$} c @{\hspace{13pt}}
    S[table-format=4.2] @{$\,\pm\,$} r 
    S[table-format=3.2] @{$\,\pm\,$} c @{\hspace{13pt}}
    S[table-format=4.2] @{$\,\pm\,$} r 
    S[table-format=3.2] @{$\,\pm\,$} c @{\hspace{13pt}}
    }
    \toprule
    \multirow{2}{*}{Methods}
    & \multicolumn{4}{c}{Hangzhou (4$\times$4)}
    & \multicolumn{4}{c}{Manhattan (16$\times$3)}
    & \multicolumn{4}{c}{Manhattan (28$\times$7)} \\
    \cmidrule(lr){2-5}\cmidrule(lr){6-9}\cmidrule(lr){10-13}
    & \multicolumn{2}{c}{Travel Time} & \multicolumn{2}{c}{Queue Length}
    & \multicolumn{2}{c}{Travel Time} & \multicolumn{2}{c}{Queue Length}
    & \multicolumn{2}{c}{Travel Time} & \multicolumn{2}{c}{Queue Length} \\
    \midrule
    Fixed-Time   & 416.88 & 0.00 & 0.96 & 0.00 & 535.60 & 0.00 & 0.89 & 0.00 & 723.08  & 0.00 & 1.17            & 0.00 \\
    SOTL         & 440.66 & 0.00 & 1.21 & 0.00 & 862.81 & 0.00 & 1.54 & 0.00 & 769.67  & 0.00 & 1.28            & 0.00 \\
    MaxPressure  & 303.31 & 0.00 & 0.29 & 0.00 & 250.14 & 0.00 & 0.14 & 0.00 & 564.36  & 0.00 & {\bfseries \phantom{00}0.70} & 0.00 \\
    \midrule
    CoLight      & 689.89 & 49.20  & 3.53 & 0.15 & 853.52  & 14.66  & 1.58 & 0.05 & 829.18  & 8.72   & 1.24 & 0.06 \\
    MPLight      & 616.67 & 204.3 & 2.63 & 1.58 & 775.45  & 116.6 & 1.50 & 0.15 & 829.73  & 4.93   & 1.44 & 0.09 \\
    AttendLight  & 508.43 & 158.6 & 1.97 & 1.35 & 712.16  & 201.2 & 1.26 & 0.39 & 736.47  & 59.58  & 1.17 & 0.05 \\
    MetaLight    & 319.56 & 13.02  & 0.53 & 0.20 & 289.73  & 127.8 & 0.34 & 0.40 & 558.45  & 28.64  & 0.79 & 0.09 \\
    E-PressLight & 708.40 & 73.76  & 3.50 & 0.14 & 786.79  & 42.53  & 1.54 & 0.08 & 819.29  & 13.20  & 1.28 & 0.10 \\
    E-MPLight    & 511.19 & 181.3 & 1.95 & 1.62 & 591.18  & 171.6 & 1.00 & 0.46 & 764.78  & 102.3 & 1.11 & 0.21 \\
    A-MPLight    & 578.57 & 138.7 & 2.71 & 1.32 & 758.68  & 160.8 & 1.31 & 0.32 & 736.50  & 82.14  & 1.24 & 0.27 \\
    UniLight     & 681.77 & 71.37  & 1.10 & 0.13 & 1303.92 & 56.34  & 1.16 & 0.05 & 1535.36 & 16.35  & 1.24 & 0.02 \\
    \midrule
    AMM w/o Mod. & 372.00 & 69.65 & 1.06 & 0.70 & 197.02 & 26.60 & 0.07            & 0.04 & 704.33 & 62.80 & 1.08 & 0.13 \\
    AMM w/o ML   & 279.95 & 3.50 & 0.15 & 0.02 & 171.34 & 3.14 & {\bfseries \phantom{00}0.03} & 0.01 & 557.65 & 23.21 & 0.75 & 0.10 \\
    AMM          & {\bfseries \phantom{0}277.95} & 3.54 & {\bfseries \phantom{00}0.14} & 0.03 & {\bfseries \phantom{0}168.36} & 2.12 & {\bfseries \phantom{00}0.03} & 0.01 & {\bfseries \phantom{0}538.93} & 6.81 & {\bfseries \phantom{00}0.70} & 0.03 \\
    \bottomrule
    \end{tabular}
    \caption{Comparison with baselines on benchmarks. Metrics are average travel time (s) and average queue length (vehicles), lower is better. Each method is run three times per benchmark and the table reports mean~$\pm$~std. Bold marks the best result in each column. Conventional controllers are deterministic under fixed flows and therefore show zero variance.}
    \label{tab:main_exp}
\end{table*}

\begin{figure*}
    \centering
    \includegraphics[width=\linewidth]{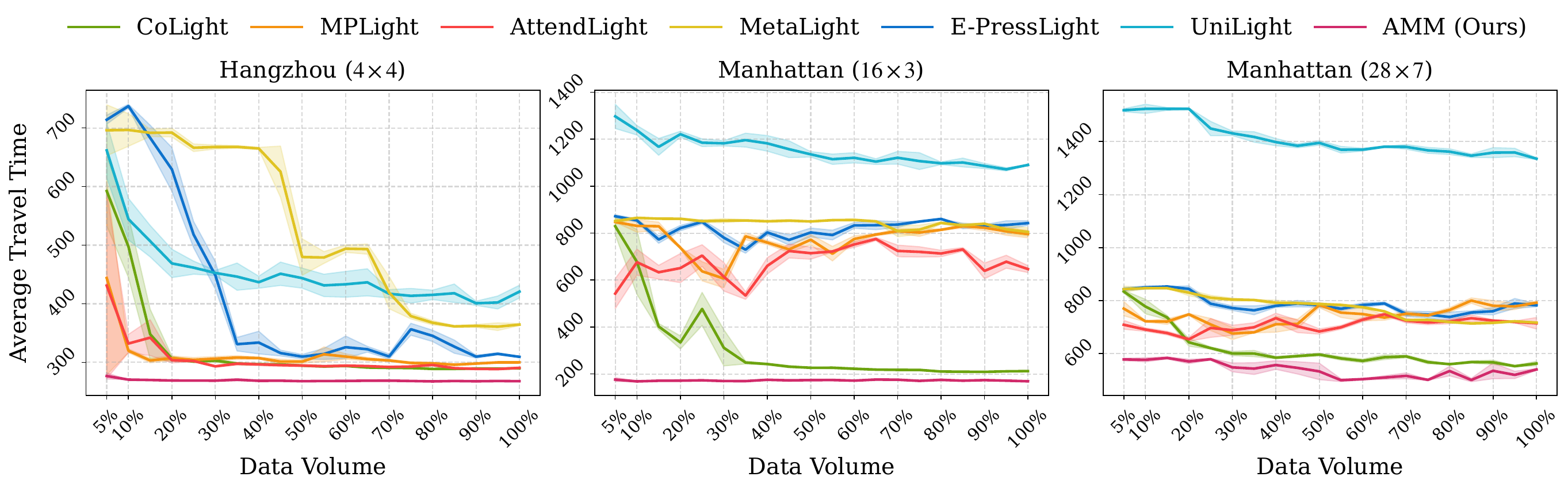}
    \caption{Data efficiency on the target domains. Average travel time as the fraction of the target interaction budget increases, where 100\% equals 60 episodes. Curves are averaged over three runs. AMM reaches strong performance with substantially less target interaction than baselines.}
    \label{fig:visualization}
\end{figure*}

\begin{figure*}
    \centering
    \includegraphics[width=\linewidth]{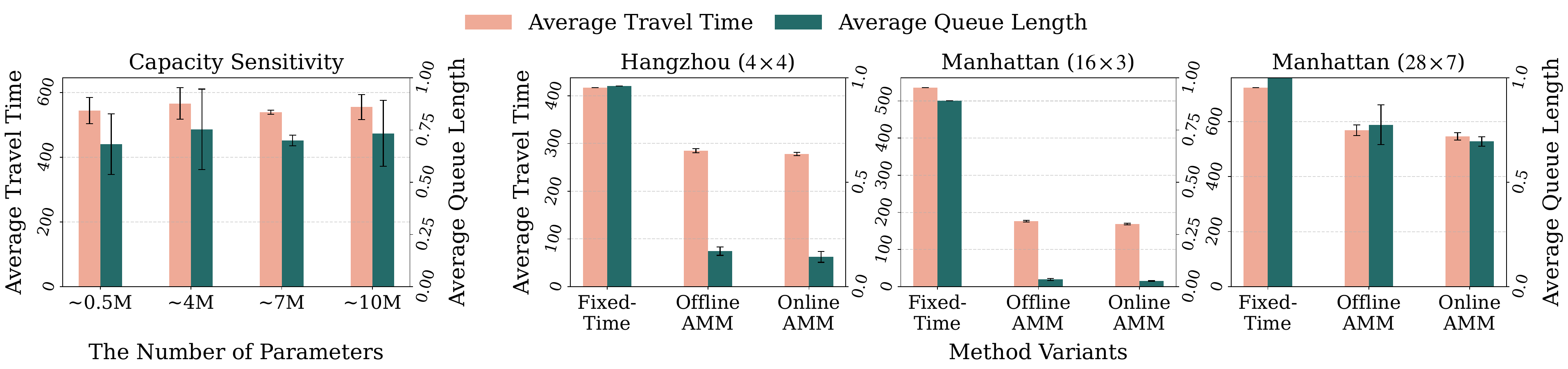}
    \caption{Analysis of AMM. \textbf{Left:} Average travel time and average queue length for AMM with different parameter counts. Performance varies little across a wide range of capacities, indicating limited sensitivity to model size under the considered budgets. \textbf{Right:} Comparison between offline and online training of the observation adapter. ``Offline AMM'' fits the adapter from non-interactive logs collected by a fixed-time controller, while ``Online AMM'' uses online interaction. Offline AMM approaches online performance with only limited logs, highlighting the low-risk adaptation enabled by the modular design.}
    \label{fig:capacity_sensitivity_and_offline_data_training}
\end{figure*}

\begin{table*}
    \centering
    \setlength{\tabcolsep}{3.5pt}
    \renewcommand{\arraystretch}{1.05}
    \begin{tabular}{
    l
    S[table-format=4.2] @{$\,\pm\,$} r @{\hspace{6pt}}
    S[table-format=4.2] @{$\,\pm\,$} c @{\hspace{15pt}}
    S[table-format=4.2] @{$\,\pm\,$} r @{\hspace{6pt}}
    S[table-format=4.2] @{$\,\pm\,$} c @{\hspace{15pt}}
    S[table-format=4.2] @{$\,\pm\,$} r @{\hspace{6pt}}
    S[table-format=4.2] @{$\,\pm\,$} c @{\hspace{15pt}}
    }
    \toprule
    \multirow{3}{*}{Source Domain}
    & \multicolumn{12}{c}{Target Domain} \\
    \cmidrule(lr){2-13}
    & \multicolumn{4}{c}{Hangzhou (4$\times$4)}
    & \multicolumn{4}{c}{Manhattan (16$\times$3)}
    & \multicolumn{4}{c}{Manhattan (28$\times$7)} \\
    \cmidrule(lr){2-5}\cmidrule(lr){6-9}\cmidrule(lr){10-13}
    & \multicolumn{2}{c}{Travel Time} & \multicolumn{2}{c}{Queue Length}
    & \multicolumn{2}{c}{Travel Time} & \multicolumn{2}{c}{Queue Length}
    & \multicolumn{2}{c}{Travel Time} & \multicolumn{2}{c}{Queue Length} \\
    \midrule
    Hangzhou (4$\times$4)
        & \multicolumn{4}{c}{/}
        & 180.67 & 1.99 & 0.05 & 0.01
        & 551.75 & 13.13 & 0.72 & 0.09 \\
    Manhattan (16$\times$3)
        & 280.57 & 2.52 & 0.16 & 0.02
        & \multicolumn{4}{c}{/}
        & 606.03 & 62.11 & 0.76 & 0.18 \\
    Manhattan (28$\times$7)
        & 280.93 & 2.57 & 0.16 & 0.02
        & 174.83 & 2.02 & 0.04 & 0.01
        & \multicolumn{4}{c}{/} \\
    \midrule
    Fixed-Time
        & 416.88 & 0.00 & 0.96 & 0.00
        & 535.60 & 0.00 & 0.89 & 0.00
        & 723.08 & 0.00 & 1.17 & 0.00 \\
    MaxPressure
        & 303.31 & 0.00 & 0.29 & 0.00
        & 250.14 & 0.00 & 0.14 & 0.00
        & 564.36 & 0.00 & 0.70 & 0.00 \\
    \bottomrule
    \end{tabular}
    \caption{Single-source pretraining study. Each row uses the source domain shown to pretrain the dynamics model and then adapts to the target domain with the standard budget. Cells report mean~$\pm$~std of average travel time and average queue length, lower is better. “/” marks the diagonal where pretraining on the same domain is not applicable. Fixed-Time and MaxPressure are included as reference non-learning controllers.}
    \label{tab:source_domain_selection}
\end{table*}

\subsubsection{Protocol}
Each episode simulates 3600 seconds.
We use a fixed control interval of 20 seconds.
At each decision step, every intersection selects one phase and holds it until the next decision step.
This results in 180 decisions per episode.
Our controller uses parameter sharing across intersections.
The same model parameters are applied to every intersection, and the controller outputs one phase action per intersection at each step.

In the transfer, we follow a leave-one-domain-out protocol.
For each target domain, the remaining two domains are used as sources.
AMM meta-learns the dynamics initialization on the source domains and then adapts to the target domain using a limited interaction budget.
We define the full budget as 60 episodes of target domain interaction. Unless otherwise stated, we report results under the full target interaction budget.
Besides, we report results under varying target data budgets, considering budgets in \{5\%, 10\%, 20\%, \dots, 100\%\}.
When comparing with learning-based baselines, we restrict their training to the same target interaction budget.
This reflects the practical setting where a new deployment provides only limited online interaction.
It also avoids conflating target performance with additional target data.

For each learning-based method, we run three seeds and report mean and standard deviation.
Conventional baselines are deterministic under fixed traffic flows and thus have zero variance in our setting.

\subsubsection{Metrics}
We report two standard metrics.
Average travel time measures the mean time a vehicle spends in the network.
Average queue length measures the mean number of waiting vehicles.
Lower values indicate better performance.

\subsection{Baselines}
\label{sec:baselines}
We compare against conventional controllers and learning-based controllers.

\paragraph{Conventional controllers.}
\textbf{Fixed-Time} \cite{allsop1971delay} cycles through phases with pre-defined green durations.
\textbf{SOTL} \cite{gershenson2004self} selects phases based on local queue pressure heuristics.
\textbf{MaxPressure} \cite{varaiya2013max} greedily selects the phase with maximum pressure and is a strong non-learning baseline.

\paragraph{Learning-based controllers.}
We include representative deep RL methods for network-level control and multi-agent coordination.
\textbf{CoLight} \cite{wei2019colight} uses graph attention to coordinate intersections.
\textbf{MPLight} \cite{chen2020toward} builds on FRAP \cite{zheng2019learning} and uses pressure-based features.
\textbf{AttendLight} \cite{oroojlooy2020attendlight} uses attention to aggregate information and learns a universal controller.
\textbf{MetaLight} \cite{zang2020metalight} applies meta-learning to improve adaptation across scenarios.
\textbf{E-PressLight} \cite{wu2021efficient} and \textbf{E-MPLight} \cite{wu2021efficient} improve pressure-based representations.
\textbf{A-MPLight} \cite{zhang2022expression} refines the pressure and demand representation.
\textbf{UniLight} \cite{jiang2022multi} uses a communication mechanism to improve coordination.

\subsection{Main Results}
\label{sec:main_results}
Table~\ref{tab:main_exp} reports the main results on average travel time and average queue length.
AMM attains the best overall performance with low variance across all three benchmarks. A consistent observation is that conventional controllers can outperform several learning-based baselines when the available training data are limited.
This is expected since conventional controllers do not require exploration.
AMM remains competitive in this regime because it transfers a dynamics model from source domains and only adapts a small number of parameters in the target domain.

\subsection{Data Efficiency Under Limited Target Interaction}
\label{sec:data_efficiency}

Figure~\ref{fig:visualization} compares average travel time as a function of target interaction budget.
AMM improves rapidly with small target budgets and reaches strong performance earlier than the baselines.
This behavior is consistent with the design of AMM.
The dynamics initialization is meta-learned from source domains, and adaptation in the target domain mainly aligns the observation adapter and fine-tunes the dynamics model.

\subsection{Ablation Studies}
\label{sec:ablation}

We include two ablations to isolate the impact of the modular design and the meta-learning procedure.
Results are shown in Table~\ref{tab:main_exp}.

\subsubsection{Effect of modularization}
The ``AMM w/o Mod." represents the non-modularized variant, which uses a single end-to-end model that directly maps observations to value estimates, and it does not separate an observation adapter from the dynamics model.
This variant performs substantially worse, particularly under observation mismatch, since it cannot reuse source data in a consistent internal space.

\subsubsection{Effect of meta-learning}
The ``AMM w/o ML" represents the sequential training variant that replaces meta learning with sequential multi-domain training.
It trains the dynamics model on source domains by standard gradient descent and then fine-tunes on the target domain.
Sequential training performs competitively but is consistently worse than AMM.
This supports the role of meta-learning in producing an initialization that adapts more effectively with limited target data.

\subsection{Further Analyses}
\label{sec:further_analyses}

\subsubsection{Sensitivity to model capacity}
We study model capacity by varying network depth and width and reporting performance as a function of parameter count.
Figure~\ref{fig:capacity_sensitivity_and_offline_data_training} left shows that AMM is relatively stable across a wide range of model sizes.
Performance saturates once the model reaches moderate capacity, and larger models do not provide consistent additional gains under the considered data budgets.

\subsubsection{Source domain selection}
To assess how source domains influence transfer, we train AMM using a single source domain and then adapt it to the target domain.
Table~\ref{tab:source_domain_selection} reports the results.
Single-source pretraining already yields reasonable performance after target adaptation, which suggests that the learned dynamics capture shared structure.
Using both source domains provides the best results, which supports aggregating multi-domain experience.

\subsubsection{Training the observation adapter with offline data}
In many deployments, logged data are easier to obtain than online interaction.
We evaluate whether the observation adapter can be trained from logged trajectories collected by a fixed-time controller.
Figure~\ref{fig:capacity_sensitivity_and_offline_data_training} right compares Offline AMM, which trains the adapter from logged data, with Online AMM, which trains with interactive target data.
Offline AMM attains competitive performance, which indicates that the modular separation enables different training strategies for different components.

\section{Conclusion, Limitations, and Future Work}
\label{sec:conclusion}

This paper studies transfer for sequential decision-making when observation interfaces differ across deployments.
We cast this setting as transferable model-based planning under observation mismatch and present AMM, a modular world‑model planner.
AMM separates a domain‑specific observation adapter from a shared internal dynamics model in a common planning state space.
The dynamics model is meta‑learned on source domains for fast target adaptation, and AMM selects actions by receding‑horizon planning with learned rollouts.

We instantiate AMM on traffic signal control with heterogeneous observation pipelines.
On CityFlow benchmarks, AMM reduces average travel time and queue length compared with conventional controllers and strong learning‑based baselines under the same target interaction budgets.
Ablations show that modularization enables reuse across observation interfaces and that meta‑learning improves adaptation over sequential multi‑domain training.
Analyses further show stable performance across model capacities, benefits from aggregating multiple source domains, and that the observation adapter trains effectively from logged data.

Several limitations remain.
The planner searches action sequences per intersection with parameter sharing rather than joint actions at network scale.
Adapter training uses simulator‑derived targets that may be unavailable or noisy in real deployments, and learned rollouts can accumulate error over the planning horizon.
Future work will explore joint or factored planning with stronger coupling, uncertainty‑aware dynamics and value estimation, weaker or self‑supervised adapter learning, and broader evaluations beyond traffic signal control.

\section{Acknowledgments}
This work was sponsored by 2025 Shanghai Key Technology Research and Development Program, Next-Generation Information Technology Project under Grant No. 25511103800.

\bibliography{reference}

@inproceedings{tobin2017domain,
  title={Domain randomization for transferring deep neural networks from simulation to the real world},
  author={Tobin, Josh and Fong, Rachel and Ray, Alex and Schneider, Jonas and Zaremba, Wojciech and Abbeel, Pieter},
  booktitle={2017 IEEE/RSJ international conference on intelligent robots and systems (IROS)},
  pages={23--30},
  year={2017},
  organization={IEEE}
}

@article{oroojlooy2020attendlight,
  title={Attendlight: Universal attention-based reinforcement learning model for traffic signal control},
  author={Oroojlooy, Afshin and Nazari, Mohammadreza and Hajinezhad, Davood and Silva, Jorge},
  journal={Advances in Neural Information Processing Systems},
  volume={33},
  pages={4079--4090},
  year={2020}
}

@article{gershenson2004self,
  title={Self-organizing traffic lights},
  author={Gershenson, Carlos},
  journal={arXiv preprint nlin/0411066},
  year={2004}
}

@article{varaiya2013max,
  title={Max pressure control of a network of signalized intersections},
  author={Varaiya, Pravin},
  journal={Transportation Research Part C: Emerging Technologies},
  volume={36},
  pages={177--195},
  year={2013},
  publisher={Elsevier}
}

@article{allsop1971delay,
  title={Delay-minimizing settings for fixed-time traffic signals at a single road junction},
  author={Allsop, Richard E},
  journal={IMA Journal of Applied Mathematics},
  volume={8},
  number={2},
  pages={164--185},
  year={1971},
  publisher={Oxford University Press}
}

@inproceedings{wei2018intellilight,
  title={Intellilight: A reinforcement learning approach for intelligent traffic light control},
  author={Wei, Hua and Zheng, Guanjie and Yao, Huaxiu and Li, Zhenhui},
  booktitle={Proceedings of the 24th ACM SIGKDD International Conference on Knowledge Discovery \& Data Mining},
  pages={2496--2505},
  year={2018}
}

@inproceedings{zheng2019learning,
  title={Learning phase competition for traffic signal control},
  author={Zheng, Guanjie and Xiong, Yuanhao and Zang, Xinshi and Feng, Jie and Wei, Hua and Zhang, Huichu and Li, Yong and Xu, Kai and Li, Zhenhui},
  booktitle={Proceedings of the 28th ACM international conference on information and knowledge management},
  pages={1963--1972},
  year={2019}
}

@inproceedings{wei2019presslight,
  title={Presslight: Learning max pressure control to coordinate traffic signals in arterial network},
  author={Wei, Hua and Chen, Chacha and Zheng, Guanjie and Wu, Kan and Gayah, Vikash and Xu, Kai and Li, Zhenhui},
  booktitle={Proceedings of the 25th ACM SIGKDD International Conference on Knowledge Discovery \& Data Mining},
  pages={1290--1298},
  year={2019}
}

@inproceedings{wei2019colight,
  title={Colight: Learning network-level cooperation for traffic signal control},
  author={Wei, Hua and Xu, Nan and Zhang, Huichu and Zheng, Guanjie and Zang, Xinshi and Chen, Chacha and Zhang, Weinan and Zhu, Yanmin and Xu, Kai and Li, Zhenhui},
  booktitle={Proceedings of the 28th ACM International Conference on Information and Knowledge Management},
  pages={1913--1922},
  year={2019}
}

@inproceedings{chen2020toward,
  title={Toward a thousand lights: Decentralized deep reinforcement learning for large-scale traffic signal control},
  author={Chen, Chacha and Wei, Hua and Xu, Nan and Zheng, Guanjie and Yang, Ming and Xiong, Yuanhao and Xu, Kai and Li, Zhenhui},
  booktitle={Proceedings of the AAAI Conference on Artificial Intelligence},
  volume={34},
  pages={3414--3421},
  year={2020}
}

@inproceedings{zang2020metalight,
  title={Metalight: Value-based meta-reinforcement learning for traffic signal control},
  author={Zang, Xinshi and Yao, Huaxiu and Zheng, Guanjie and Xu, Nan and Xu, Kai and Li, Zhenhui},
  booktitle={Proceedings of the AAAI Conference on Artificial Intelligence},
  volume={34},
  pages={1153--1160},
  year={2020}
}

@inproceedings{zhang2022expression,
  title={Expression might be enough: representing pressure and demand for reinforcement learning based traffic signal control},
  author={Zhang, Liang and Wu, Qiang and Shen, Jun and L{\"u}, Linyuan and Du, Bo and Wu, Jianqing},
  booktitle={International Conference on Machine Learning},
  pages={26645--26654},
  year={2022},
  organization={PMLR}
}

@article{jiang2022multi,
  title={Multi-agent reinforcement learning for traffic signal control through universal communication method},
  author={Jiang, Qize and Qin, Minhao and Shi, Shengmin and Sun, Weiwei and Zheng, Baihua},
  journal={arXiv preprint arXiv:2204.12190},
  year={2022}
}

@inproceedings{zhang2019cityflow,
  title={Cityflow: A multi-agent reinforcement learning environment for large scale city traffic scenario},
  author={Zhang, Huichu and Feng, Siyuan and Liu, Chang and Ding, Yaoyao and Zhu, Yichen and Zhou, Zihan and Zhang, Weinan and Yu, Yong and Jin, Haiming and Li, Zhenhui},
  booktitle={The world wide web conference},
  pages={3620--3624},
  year={2019}
}

@article{wei2019survey,
  title={A Survey on Traffic Signal Control Methods},
  author={Wei, Hua and Zheng, Guanjie and Gayah, Vikash and Li, Zhenhui},
  journal={arXiv preprint arXiv:1904.08117},
  year={2019}
}

@article{wu2021efficient,
  title={Efficient pressure: Improving efficiency for signalized intersections},
  author={Wu, Qiang and Zhang, Liang and Shen, Jun and L{\"u}, Linyuan and Du, Bo and Wu, Jianqing},
  journal={arXiv preprint arXiv:2112.02336},
  year={2021}
}

@inproceedings{finn2017model,
  title={Model-agnostic meta-learning for fast adaptation of deep networks},
  author={Finn, Chelsea and Abbeel, Pieter and Levine, Sergey},
  booktitle={International conference on machine learning},
  pages={1126--1135},
  year={2017},
  organization={PMLR}
}

@book{ghallab2016automated,
  title={Automated planning and acting},
  author={Ghallab, Malik and Nau, Dana and Traverso, Paolo},
  year={2016},
  publisher={Cambridge University Press}
}

@inproceedings{sun2022transferobs,
  title  = {Transfer RL across Observation Feature Spaces via Model-Based Regularization},
  author = {Sun, Yanchao and Zheng, Ruijie and Wang, Xiyao and Cohen, Andrew and Huang, Furong},
  year   = {2022},
  booktitle = {International Conference on Learning Representations}
}

@inproceedings{xing2021domain,
  title={Domain adaptation in reinforcement learning via latent unified state representation},
  author={Xing, Jinwei and Nagata, Takashi and Chen, Kexin and Zou, Xinyun and Neftci, Emre and Krichmar, Jeffrey L},
  booktitle={Proceedings of the AAAI Conference on Artificial Intelligence},
  volume={35},
  pages={10452--10459},
  year={2021}
}

@inproceedings{zhao2020sim,
  title={Sim-to-real transfer in deep reinforcement learning for robotics: a survey},
  author={Zhao, Wenshuai and Queralta, Jorge Pe{\~n}a and Westerlund, Tomi},
  booktitle={2020 IEEE symposium series on computational intelligence (SSCI)},
  pages={737--744},
  year={2020},
  organization={IEEE}
}

@article{garcia2015comprehensive,
  title={A comprehensive survey on safe reinforcement learning},
  author={Garc{\i}a, Javier and Fern{\'a}ndez, Fernando},
  journal={Journal of Machine Learning Research},
  volume={16},
  number={1},
  pages={1437--1480},
  year={2015}
}

@article{levine2020offline,
  title={Offline reinforcement learning: Tutorial, review, and perspectives on open problems},
  author={Levine, Sergey and Kumar, Aviral and Tucker, George and Fu, Justin},
  journal={arXiv preprint arXiv:2005.01643},
  year={2020}
}

@article{mayne2000constrained,
  title={Constrained model predictive control: Stability and optimality},
  author={Mayne, David Q and Rawlings, James B and Rao, Christopher V and Scokaert, Pierre OM},
  journal={Automatica},
  volume={36},
  number={6},
  pages={789--814},
  year={2000},
  publisher={Elsevier}
}

@article{rawlings2017model,
  title={Model predictive control: theory, computation, and design, vol. 2},
  author={Rawlings, James Blake and Mayne, David Q and Diehl, Moritz and others},
  journal={Madison, WI: Nob Hill Publishing},
  year={2017}
}

@article{chua2018deep,
  title={Deep reinforcement learning in a handful of trials using probabilistic dynamics models},
  author={Chua, Kurtland and Calandra, Roberto and McAllister, Rowan and Levine, Sergey},
  journal={Advances in neural information processing systems},
  volume={31},
  year={2018}
}

@article{ha2018world,
  title={World models},
  author={Ha, David and Schmidhuber, J{\"u}rgen},
  journal={arXiv preprint arXiv:1803.10122},
  volume={2},
  number={3},
  year={2018}
}

@inproceedings{hafner2019learning,
  title={Learning latent dynamics for planning from pixels},
  author={Hafner, Danijar and Lillicrap, Timothy and Fischer, Ian and Villegas, Ruben and Ha, David and Lee, Honglak and Davidson, James},
  booktitle={International conference on machine learning},
  pages={2555--2565},
  year={2019},
  organization={PMLR}
}

@article{hafner2019dream,
  title={Dream to control: Learning behaviors by latent imagination},
  author={Hafner, Danijar and Lillicrap, Timothy and Ba, Jimmy and Norouzi, Mohammad},
  journal={arXiv preprint arXiv:1912.01603},
  year={2019}
}

@article{schrittwieser2020mastering,
  title={Mastering atari, go, chess and shogi by planning with a learned model},
  author={Schrittwieser, Julian and Antonoglou, Ioannis and Hubert, Thomas and Simonyan, Karen and Sifre, Laurent and Schmitt, Simon and Guez, Arthur and Lockhart, Edward and Hassabis, Demis and Graepel, Thore and others},
  journal={Nature},
  volume={588},
  number={7839},
  pages={604--609},
  year={2020},
  publisher={Nature Publishing Group UK London}
}

@inproceedings{sun2024crosslight,
  title={Crosslight: Offline-to-online reinforcement learning for cross-city traffic signal control},
  author={Sun, Qian and Zha, Rui and Zhang, Le and Zhou, Jingbo and Mei, Yu and Li, Zhiling and Xiong, Hui},
  booktitle={Proceedings of the 30th ACM SIGKDD Conference on Knowledge Discovery and Data Mining},
  pages={2765--2774},
  year={2024}
}

@inproceedings{jiang2024x,
  title={X-light: Cross-city traffic signal control using transformer on transformer as meta multi-agent reinforcement learner},
  author={Jiang, Haoyuan and Li, Ziyue and Wei, Hua and Xiong, Xuantang and Ruan, Jingqing and Lu, Jiaming and Mao, Hangyu and Zhao, Rui},
  booktitle={International Joint Conferences on Artificial Intelligence},
  year={2024}
}

@article{mei2024libsignal,
  title={Libsignal: An open library for traffic signal control},
  author={Mei, Hao and Lei, Xiaoliang and Da, Longchao and Shi, Bin and Wei, Hua},
  journal={Machine Learning},
  volume={113},
  number={8},
  pages={5235--5271},
  year={2024},
  publisher={Springer}
}

@article{zhang2023datalight,
  title={DataLight: Offline Data-Driven Traffic Signal Control},
  author={Zhang, Liang and Zhang, Yutong and Deng, Jianming and Li, Chen},
  journal={arXiv preprint arXiv:2303.10828},
  year={2023}
}

@inproceedings{bokade2025offlight,
  title={OffLight: An Offline Multi-Agent Reinforcement Learning Framework for Traffic Signal Control},
  author={Bokade, Rohit and Jin, Xiaoning},
  booktitle={2025 IEEE 21st International Conference on Automation Science and Engineering (CASE)},
  pages={2730--2737},
  year={2025},
  organization={IEEE}
}

@inproceedings{du2023safelight,
  title={Safelight: A reinforcement learning method toward collision-free traffic signal control},
  author={Du, Wenlu and Ye, Junyi and Gu, Jingyi and Li, Jing and Wei, Hua and Wang, Guiling},
  booktitle={Proceedings of the AAAI conference on artificial intelligence},
  volume={37},
  pages={14801--14810},
  year={2023}
}

@article{mei2023reinforcement,
  title={Reinforcement learning approaches for traffic signal control under missing data},
  author={Mei, Hao and Li, Junxian and Shi, Bin and Wei, Hua},
  journal={arXiv preprint arXiv:2304.10722},
  year={2023}
}

\end{document}